\pdfoutput=1

\documentclass[11pt]{article}

\usepackage{ACL2023}

\usepackage{times}
\usepackage{latexsym}

\usepackage[normalem]{ulem}
\usepackage{enumitem}
\usepackage{graphicx}
\usepackage{amsmath}
\usepackage{amssymb}
\usepackage{algorithm}
\usepackage[noend]{algpseudocode}
\usepackage{caption}
\usepackage{subcaption}
\usepackage{float}
\usepackage{colortbl}

\usepackage[T1]{fontenc}

\usepackage[utf8]{inputenc}

\usepackage{microtype}

\usepackage{inconsolata}

\title{Yes, this Way! Learning to Ground Referring Expressions into Actions\\ with Intra-episodic Feedback from Supportive Teachers}

\author{Philipp Sadler$^\mathbf{1}${\normalfont ,} Sherzod Hakimov$^\mathbf{1}$ {\normalfont and} David Schlangen$^\mathbf{1,2}$ \\
  $^\mathbf{1}$CoLabPotsdam / Computational Linguistics \\
  Department of Linguistics, University of Potsdam, Germany \\
  $^\mathbf{2}$German Research Center for Artificial Intelligence (DFKI), Berlin, Germany \\
  \texttt{firstname.lastname@uni-potsdam.de}\\}

\begin{document}
\maketitle
\begin{abstract}
The ability to pick up on language signals in an ongoing interaction is crucial for future machine learning models to collaborate and interact with humans naturally. In this paper, we present an initial study that evaluates intra-episodic feedback given in a collaborative setting. We use a referential language game as a controllable example of a task-oriented collaborative joint activity. A teacher utters a referring expression generated by a well-known symbolic algorithm (the ``Incremental Algorithm'') as an initial instruction and then monitors the follower's actions to possibly intervene with intra-episodic feedback (which does not explicitly have to be requested). 
We frame this task as a reinforcement learning problem with sparse rewards and learn a follower policy for a heuristic teacher. Our results show that intra-episodic feedback allows the follower to generalize on aspects of scene complexity and performs better than providing only the initial statement.

\end{abstract}

\section{Introduction} %

The communicative acts of humans in collaborative situations can be described as two parts of a joint act: signalling and recognizing. In such joint activities, these signals work as coordination devices to increment on the current common ground of the participants \citep{clark_using_1996}. The ability to act on these language signals is crucial for future machine learning models to naturally collaborate and interact with humans \citep{lemon_conversational_2022,Fernandez-2011}. 
Such a collaborative interaction with humans usually happens fluently, where one communicative act is performed after the other. 
The framework of reinforcement learning (RL) \citep{Sutton1998} describes such mechanics where an agent is exposed in steps to observations of an environment with dynamic factors such as the position of objects or 
language expressions. The goal is that the agent learns to behave generally well in a particular environment solely based on the observations it makes and rewards it gets.

\begin{figure}[t]
    \begin{center}
        \includegraphics[width=0.45\textwidth]{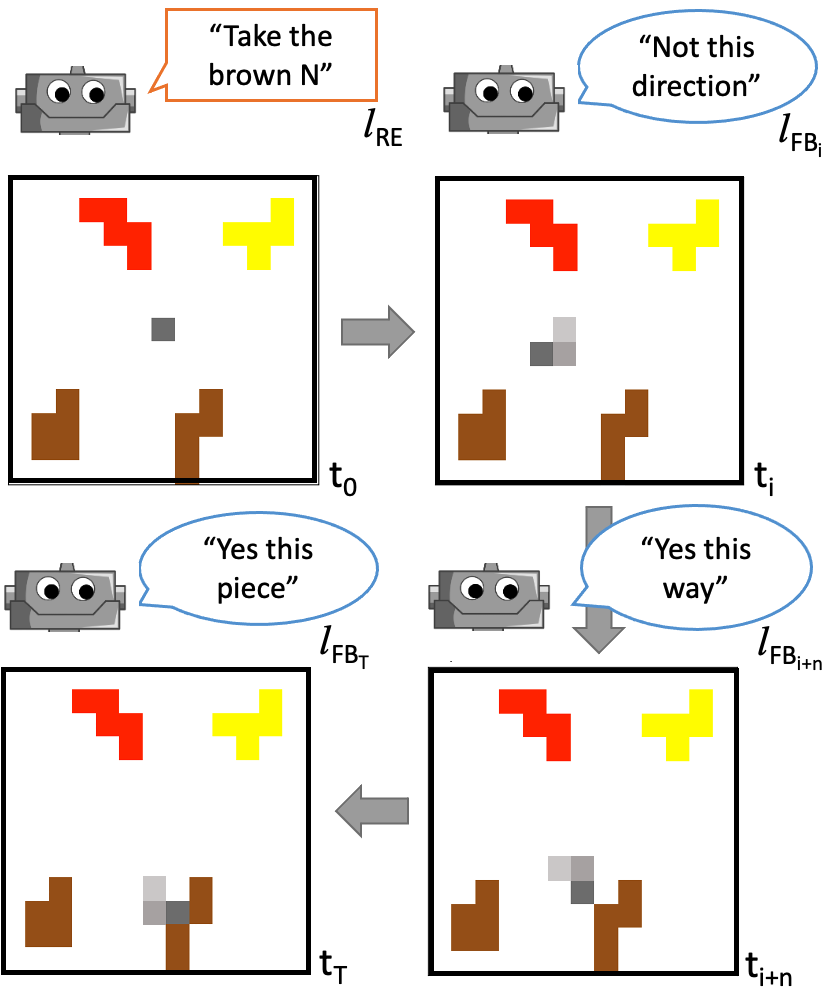}
        \vspace{-0.3cm}
    \end{center}
    \caption{An exemplary interaction between a teacher and a follower that controls the gripper (the grey square). After an initial referring expression $l_{\text{RE}}$ at $t_0$, the teacher provides feedback $l_{\text{FB}_t}$ based on the follower's actions until the correct piece is selected. %
    }
    \label{fig:example_board}
   \vspace{-0.5cm}
\end{figure}

A key challenge here is the variability of expressions in language that can be said to the agent during an interaction. Even in relatively simple environments, there might arise an overwhelming amount of situations for an agent to handle~ \citep{chevalier-boisvert_babyai_2019}.
Recent work on collaborative agents focuses on large pre-collected datasets for imitation learning to learn agents in complex simulated visual environments \citep{gao_dialfred_2022,padmakumar_teach_2022,pashevich_episodic_2021} or frames the learning as a contextual bandit problem \citep{suhr_continual_2022,suhr_executing_2019}. Nevertheless, other work has shown that intermediate language inputs are a valuable signal to improve the agent's learning performance in task-oriented visual environments \citep{co-reyes_guiding_2019,mu_improving_2022}. %

In this paper, we present an initial study that evaluates a follower's learning success given a teacher's intra-episodic feedback in a collaborative setting. We use a referential language game (in English) as a controllable example of a task-oriented collaborative joint activity (see Figure \ref{fig:example_board}).  %
In this game one player (the follower) is supposed to select a piece based on the another player's directives (the teacher). We assume a teacher that utters referring expressions as initial instructions and then responds to the follower's actions with intra-episodic feedback. We frame this as a RL problem with sparse rewards where the intermediate feedback is not part of the reward function but its potential usefulness is learnt by the follower alone.\footnote{Code is publicly available at \url{https://github.com/clp-research/intra-episodic-feedback}.}%

\section{Related Work} %

\paragraph{Vision and language navigation.} In vision and language navigation, an agent is given a natural language instruction which is to be understood to navigate to the correct goal location in a visually observed environment \citep{gu_vision-and-language_2022}. The follower can usually ask an Oracle for further information, if necessary \citep{DBLP:conf/cvpr/NguyenDBD19,DBLP:conf/emnlp/NguyenD19,DBLP:conf/nips/FriedHCRAMBSKD18}. We extend on this idea and aim for an ongoing interaction with corrections that loosens the turn-based paradigm by letting the Oracle choose when to speak as part of the environment. Hence, in our reference game, the language back-channel for the follower is cut, so that we force the follower to rely more on the visual observations for task success.

\paragraph{Continual learning from human feedback.} \citet{suhr_continual_2022} let humans instruct the follower and then ask them to rate the agent's behaviour (thumbs up or down). This binary feedback is used for further training as the reward signal in a contextual bandit framework. They show that the agent improves over several interactions with humans. 
Similarly we evaluate the learning process in the context of RL because it imposes ''weaker constraints on the regularity of the solution`` \citep{DBLP:conf/cvpr/NguyenDBD19}, but take a broadly available, off-the-shelf learning algorithm \citep{schulman_proximal_2017} to directly study the effects of different kinds of feedback. The feedback given to our agent is of natural language and not directly bound to the reward; the follower needs to learn the meaning of the language feedback itself. 

\paragraph{Language-guided policy learning.} \citet{chevalier-boisvert_babyai_2019} compared the sampling complexity of RL and imitation learning (IL) agents on various language-conditioned tasks. They proposed a 2-dimensional visual environment called \textit{Minigrid} in which an agent is given a single mission statement that instructs the agent to achieve a specific state, e.g. ''Take the red ball``. In contrast to them we intentionally do not use IL approaches, because then the agent would have already learnt how to ground the language signals. We want to test if the agent can pick-up on the language from the interaction alone. For this, we similarly propose a diagnostic environment to directly control for the distributions of target objects (cf. skewed distribution of target objects in CVDN \citep{DBLP:journals/corr/abs-1907-04957}) and feedback signals.

Other work uses the \textit{Minigrid} environment to propose a meta-training approach that improves the learning via natural language corrections, e.g.\ ``Pick up the green ball'' \citep{co-reyes_guiding_2019}. The agent is given an episodic correction if a specific task cannot be solved. In this way, the agent must not only ground the mission statement but also ground the corrections into actions. \citet{mu_improving_2022} improve policy learning with intra-episodic natural language sub-goals e.g.\ ``Pick up the ball''. These sub-goals are provided by a trained teacher policy when a previous sub-goal has been reached. In contrast, we rather follow earlier work  \citep{engonopoulos_predicting_2013} on monitoring execution and use a heuristic teacher which provides intra-episodic language feedback whenever it appears feasible. The agent has to learn that certain pairs of feedback and behaviour at a specific time-step lead to the task's success and others to failure.

\section{The CoGRIP environment} %
We use a \textbf{Co}llaborative \textbf{G}ame of \textbf{R}eferential and \textbf{I}nteractive language with \textbf{P}entomino pieces as a controllable setting. A teacher instructs a follower to select a specific piece using a gripper. Both are constrained as follows: The teacher can provide utterances but cannot move the gripper. The follower can move the gripper but is not allowed to provide an utterance. This asymmetry in knowledge and skill forces them to work together and coordinate. \citet{zarries_pentoref_2016} found that this settings leads to diverse language use on the teacher's side.

\begin{figure*}[t]
    \begin{center}
        \includegraphics[width=0.9\textwidth]{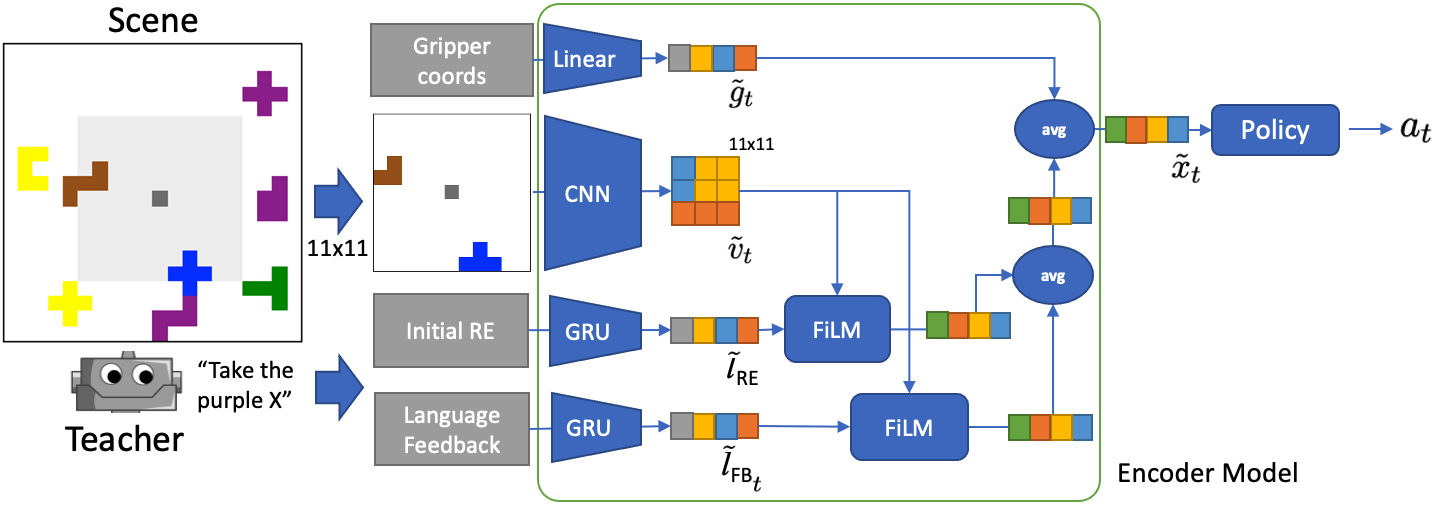}
    \end{center}
    \caption{The information flow through our encoder model which produces the features $\tilde{x}_t$ as an input to the policy.}
    \label{fig:model}
   \vspace{-0.5cm}
\end{figure*}

\subsection{Problem Formulation}
The follower has to navigate a gripper to select a piece described by the teacher. We frame this task as a RL problem with sparse rewards. At each time-step $t$, given an observation $o_t \in \mathcal{O}$ of the environment, the agent has to select an action $a_t \in \{\texttt{LEFT, RIGHT, UP, DOWN, WAIT, GRIP}\}$ such that the overall resulting sequence of actions $(a_0,...,a_t,...,a_T)$ maximizes the sparse reward $\mathcal{R}(o_T)=r$. An episode ends when the \texttt{GRIP} action is chosen, and the gripper position $g_t$ is in the boundaries of a piece. An episode also ends when $t$ reaches $T_{max}=100$. Following \citet{chevalier-boisvert_babyai_2019}, the reward function returns a basic reward minus the movement effort $\mathcal{R} = 1 - 0.9 * (T / T_{max})$. We extend this formulation and give an additional bonus of $+1$ if the correct piece has been taken or a penalty of $-1$ when the wrong or no piece has been taken at all.

\subsection{Environment}

The environment exposes at each time-step $t$ an observation $o_t$ that contains the gripper coordinates $g_t = (x,y)$, the initial referring expression $l_{\text{RE}}$, the language feedback $l_{\text{FB}_t}$ (which might be empty) and a partial view $v_t$ of the scene. While the scene as a whole is represented as a 2-dimensional image (with RGB colour channel), %
the partial view represents a $11\times11$-sized cut out, centered on the gripper position (see Figure~\ref{fig:model}). The teacher generates the initial and feedback statements.

\subsection{Teacher}

For the teacher, we assume a heuristic behaviour (a fix policy) that has been shown to lead to collaborative success with humans \citep{Goetze-2022} and leave the complexity of learning in a multi-agent setting \citep{DBLP:journals/air/GronauerD22} for future work. The teacher produces an initial referring expression $l_{\text{RE}}=(w_0,...,w_N)$ where $N$ is the message length and $w_i$ is a word in the vocabulary. The production rule is implemented following the Incremental Algorithm (\textsc{ia}) \citep{dale_computational_1995} that is given the symbolic representations of the pieces on the board (see Appendix \ref{appendix:teacher}). The teacher provides a feedback message $l_{\text{FB}_t}=(w_0,...,w_N)$ at a time-step $t_{>0}$ when the gripper's position $g_t$ has exceeded a pre-defined distance threshold $D_{\text{dist}}=3$ compared to the gripper's last position of feedback $g_{\text{FB}_{\text{last}}}$ or it is over a piece. The generated feedback is of positive sentiment (``Yes this way/piece'') when the gripper  is then closer to or over the target piece and negative otherwise (``Not this direction/piece''). Alternatively, suppose the follower does not exceed the distance threshold after $D_{\text{time}}=6$ time-steps the feedback message is the same as the initial statement. %
Overall, the property values and sentence templates lead to a small vocabulary of 33 words.%

\subsection{Follower} 

The follower agent has to move the gripper and successfully grip a piece solely based on the observations. The observations $o_t=(v_t,g_t,l_{\text{RE}},l_{\text{FB}_t})$ are mapped to 128-dimensional features $\tilde{x}_t \in \mathbb{R}$ using the encoder model (see Figure~\ref{fig:model}). Following \citet{chevalier-boisvert_babyai_2019}, the word embeddings (which are learned from scratch) of the language inputs are fed through a Gated Recurrent Unit (GRU)~\citep{cho-etal-2014-learning} and then combined with the embedded visual features using a Feature-wise Linear Modulation (FiLM) layer~\citep{DBLP:conf/aaai/PerezSVDC18}. These language conditioned visual features are then max pooled, averaged and  again averaged with the gripper position. Given the resulting features $\tilde{x}_t$, we learn a parameterised policy $\pi(\tilde{x}_t;\theta) \sim a_t$ that predicts a distribution over the action space. We use the Proximal Policy Optimization (PPO)~\citep{schulman_proximal_2017} implementation of \textit{StableBaselines3} v1.6.2 \citep{stable-baselines3} to train the policy in our environment.

\subsection{Tasks} %

The follower has to grip an intended target piece among several other pieces (the distractors). Thus a task is defined by the number of pieces, the target piece and the map size. The pieces for the tasks are instantiated from symbolic representations: a tuple of shape (9), color (6) and position (8) which leads to $432$ possible piece symbols. For our experiments we use all of these symbols as targets, but split them into distinct sets (Appendidx~\ref{appendix:data_generation}). Therefore the targets for testing tasks are distinct from the ones in the training tasks. We ensure the reproducibility of our experiments by constructing $3300$ training, $300$ validation, $720$ testing tasks representing scenes with a map size of $20\times20$ and 4 or 8 pieces.

\section{Experiments} %

In this section we explore the effects of the teacher's language and intra-episodic feedback on the follower's success and ask whether the follower generalizes on aspects of scene complexity.

\subsection{Which referential language is most beneficial for the agent's learning success?}

\begin{figure}[t]
    \begin{center}
        \includegraphics[width=0.45\textwidth]{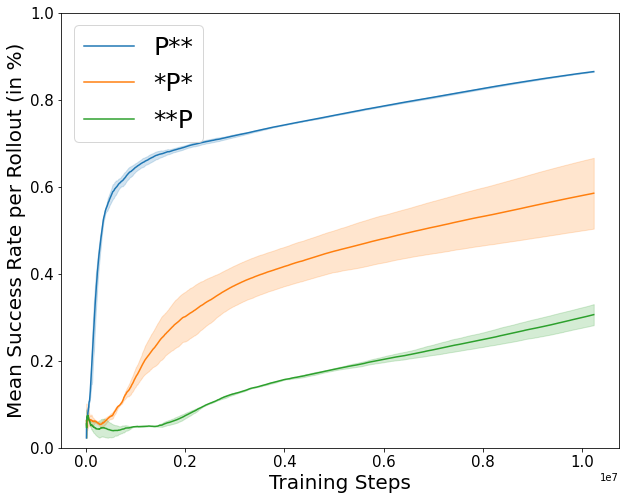}
        \vspace{-0.5cm}
    \end{center}
    \caption{The mean success rates per rollout (during training) grouped by the teacher's preference of position i.e. teacher's with \texttt{P**} start with the position description to rule out distractors and teacher's with \texttt{**P} use the position only, when color or shape are not enough to distinguish the target piece from others. The curves show that a preference for position descriptions lead to faster training success of the follower.}
    \label{fig:preference_order}
\end{figure}

As suggested by \citet{Madureira-2020} we explore the question of which language is most effective. The \textsc{ia} constructs the initial reference by following a preference order over object properties \citep{krahmer_is_2012}. We hypothesize that a particular order might be more or less suitable depending on the task. Thus we conduct a series of experiments \textit{without} the feedback signal where the preference order is varied as the permutation of color, shape and position. Our results indicate that such orders perform better that prioritize to mention positional attributes as distinguishing factors of the target piece (see Table~\ref{table:results}). This is reasonable as the directional hint reduces the agent's burden for broader exploration. The follower is able to pick up early on these positional clues and performs overall better during training (see Figure~\ref{fig:preference_order}).

\begin{table}[t]
\vspace{0.2cm}
\centering
    \begin{small}
    \begin{tabular}{|l|l|l|l|l|} 
    \hline
               & \multicolumn{2}{c|}{only initial RE} & \multicolumn{2}{c|}{with intra-episodic Feedback}  \\ 
    Pr.Or.     &   mSR & mEPL                     & mSR             & mEPL  \\ 
    \hline
    \texttt{C-P-S}      & 39.44 & 59.71                    & 87.64 (+48.2)  & 22.12 (-37.6)   \\
    \texttt{S-P-C}      & 25.42 & 68.49                    & 78.75 (+53.3)  & 32.44 (-36.0)   \\ 
    \hline
    \texttt{P-C-S}      & \textbf{73.19} & \textbf{23.06}  & 93.89 (+20.7)  & 15.39 (-7.7)    \\ 
    \texttt{P-S-C}      & 70.56 & 23.10                    & \textbf{94.44} (+23.9)  & \textbf{14.80} (-8.3)    \\ 
    \hline
    \texttt{C-S-P}      & 15.14 & 81.18                    & 79.86 (+64.7)  & 33.06 (-48.1)   \\
    \texttt{S-C-P}      & 13.33 & 85.30                    & 70.69 (+57.4)  & 42.71 (-42.6)   \\
    \hline
    \end{tabular}
    \end{small}
    \vspace{-0.2cm}
    \caption{The mean success rates (mSR in \%) and episodes lengths (mEPL) of the agent on the test tasks when the teacher follows a particular preference order over target piece properties (color (\texttt{C}), shape (\texttt{S}), position (\texttt{P})) with language feedback and without it (only initial RE). A shortest path solver reaches $10.96$ mEPL.}
    \label{table:results}
\end{table}

\begin{table}[!t]
\vspace{0.2cm}
\centering
    \begin{small}
    \begin{tabular}{|l|c|c|c|c|} 
    \hline
    Generalization Tasks & \texttt{P-C-S} & w/ FB & \# Tasks \\ 
    \hline
    Test \texttt{30x30} (\texttt{12P},\texttt{18P}) & 39.17   & 80.56 & 360           \\ 
    Test \texttt{30x30} (\texttt{4P},\texttt{8P})   & 61.94   & 91.39 & 360           \\ 
    Holdout \texttt{20x20}   (\texttt{4P},\texttt{8P})     & 63.31   & 94.44 & 864           \\
    \hline
    \hline
    Test \texttt{20x20} (\texttt{4P},\texttt{8P})   & 73.19   & 93.89 & 720\\
    \hline
    \end{tabular}
    \end{small}
    \vspace{-0.2cm}
    \caption{The mean success rates (mSR in \%) of the best agent (a teacher with pref. order \texttt{P-C-S}) on the generalization tasks. The agent with the intra-episodic feedback (w/ FB) performs much better on these more complex scenes. Number of pieces abbreviated, for example \texttt{4P} means 4 pieces. Map sizes given by \texttt{NNxNN}.}
    \label{table:generalize}
\end{table}

\subsection{What is the agent's performance gain with intra-episodic feedback in our setting?}

We conduct the same experiments as above \textit{with} intra-episodic language feedback to measure its effect on the follower's success rate. Our results show that the follower achieves higher success rates with intra-episodic feedback among all preference orders (see Table~\ref{table:results}). We also notice that the gain is higher for the low-performing preference orders. This shows that the intra-episodic feedback is a valuable signal for the follower to overcome missing directives in the initial referring expressions. The agent can learn strategies incorporating the feedback signals. This is an interesting finding because language feedback is not part of the reward function and could be empty.%

\subsection{Does intra-episodic feedback help the agent to generalize on scene complexity?}

As a proxy for generalization capabilities, we take the best performing follower and raise the complexity of the \textit{testing} scenes along two dimensions (i) we increase the map size to $30\times30$ and (ii) put up to $18$ pieces on the board. In addition, we hold out $72$ combinations of piece shapes and colors that have never been seen during training. Our results show that the agent trained with intra-episodic feedback is able to perform better (i) on the larger map size, (ii) the higher number of pieces and (iii) the new target pieces compared to the one without (see Table~\ref{table:generalize}).

\section{Conclusion}

In this work, we studied the effects of a teacher's language and intermediate interventions (the feedback) towards a learner's success and whether the learner generalizes on aspects of scene complexity.
Our results show that there is a most beneficial language for the teacher. Its intra-episodic feedback allows the learner to learn faster and generalize better than without intermediate help. An exciting direction for further work is to show the benefits of language feedback for other reinforcement learning problems, to overcome the limits of the heuristic teacher strategy and to reduce the need for feedback after successful training.

\section{Limitations}

\paragraph{Limits on visual variability and naturalness.} The Pentomino domain can only serve as an abstraction for referring expression generations in visual domains. The amount of objects is limited to 9 different shapes and the number of colors is reduced to 6 as well. The positions are chosen to be discrete and absolute while real-world references might include spatial relations. Furthermore, the pieces show no texture or naturalness, but are drawn with a solid color fill. We choose this simplified domain to focus on the interaction between the follower and the teacher and left the evaluation of the proposed models on more realistic looking scenes for further work. Nevertheless, we think our approach can also be applied to photo-realistic environments \citep{DBLP:conf/nips/RamakrishnanGWM21,DBLP:journals/corr/abs-1712-05474}.

\paragraph{Limits on variability of the referring expressions.} We only explored expressions that are generate by the Incremental Algorithm. Moreover, we choose a fix property value order (color is mentioned before shape is mentioned before position) for the realisation of the template's surface structure and left the exploration for a higher variability to further work.

\paragraph{Limits on variability of the feedback signal.} In this work we used a heuristic teacher with a fixed behavior to provide the intermediate feedback to the follower. We choose this Oracle speaker for better control over the experiments and to focus on the research questions of which feedback is most helpful and how it should be presented (contain which information). We are aware that in natural interaction the teacher's responses might be more dynamic and can be potentially learnt in a much more complex multi-agent RL settings which would go beyond our focused contribution here. Still this is an interesting prospect for future research.

\section{Ethics Statement}
For now, we see no immediate threats regarding this work, because the experiments are performed in a controlled setting of an abstract domain. But since this research has collaborative agents in prospect people might use more advanced stages of this technique to train agents on possibly other tasks. Thus we encourage everyone to apply such a technology only for good use and to avoid harmful applications.

\section*{Acknowledgements} We want to thank the anonymous reviewers for their comments. This work was funded by the \textit{Deutsche Forschungsgemeinschaft} (DFG, German Research Foundation) – 423217434 (``RECOLAGE'') grant.

\bibliography{anthology,custom}
\bibliographystyle{acl_natbib}

\newpage 
\appendix

\section{Appendix}

\subsection{Teacher Details}
\label{appendix:teacher}

\paragraph{Hyperparameters}
\begin{itemize}
    \itemsep0em
    \item $D_{\text{dist}} = 3$
    \item $D_{\text{time}} = 6$
    \item preference\_order
\end{itemize}
The distances between two coordinates $(p_1,p_2)$ are calculated as the euclidean distance.
\paragraph{The Incremental Algorithm (\textsc{ia})}

The Algorithm~\ref{alg:ia}, in the formulation of \cite{dale_computational_1995},
is supposed to find the properties that uniquely identify an object among others given a preference over properties. To accomplish this the algorithm is given the property values $\mathcal{P}$ of distractors in $M$ and of a referent $r$. Then the algorithm excludes distractors in several iterations until either $M$ is empty or every property of $r$ has been tested. During the exclusion process the algorithm computes the set of distractors that do \textit{not} share a given property with the referent and stores the property in $\mathcal{D}$. These properties in $\mathcal{D}$ are the ones that distinguish the referent from the others and thus will be returned.

The algorithm has a meta-parameter $\mathcal{O}$, indicating the \textit{preference order}, which determines the order in which the properties of the referent are tested against the distractors. In our domain, for example, when \textit{color} is the most preferred property, the algorithm might return \textsc{blue}, if this property already excludes all distractors. When \textit{shape} is the preferred property and all distractors do \textit{not} share the shape \textsc{T} with the referent, \textsc{T} would be returned. Hence even when the referent and distractor pieces are the same, different preference orders might lead to different expressions.

\begin{algorithm}[h]
\caption{The \textsc{ia} on symbolic properties as based on the formulation by \citet{deemter_2016}}\label{alg:ia}
\begin{algorithmic}[1]
\Require{A set of distractors $M$, a set of property values $\mathcal{P}$ of a referent $r$ and a linear preference order $\mathcal{O}$ over the property values $\mathcal{P}$}
\State{$\mathcal{D} \gets \emptyset $}

\For{$P$ in $\mathcal{O}(\mathcal{P})$}
{
\State{$\mathcal{E} \gets \{ m \in M: \neg P(m)$\}}
\If{$\mathcal{E} \ne \emptyset$}
    \State Add $P$ to $\mathcal{D}$
    \State Remove $\mathcal{E}$ from $M$
\EndIf
\EndFor
}
\State{\textbf{return} $\mathcal{D}$}
\end{algorithmic}
\end{algorithm}

\newpage

\paragraph{Referring Expression Templates}
\label{appendix:templates}

There are 3 expression templates that are used when only a single property value of the target piece is returned by the Incremental Algorithm (\textsc{ia}): 
\begin{itemize}
    \itemsep0em
    \item \textit{Take the [color] piece}
    \item \textit{Take the [shape]}
    \item \textit{Take the piece at [position]}
\end{itemize}
Then there are 3 expression templates that are selected when two properties are returned:
\begin{itemize}
    \itemsep0em
    \item \textit{Take the [color] [shape]}
    \item \textit{Take the [color] piece at [position]}
    \item \textit{Take the [shape] at [position]}
\end{itemize}
And finally there is one expression templates that lists all property values to identify a target piece:
\begin{itemize}
    \itemsep0em
    \item \textit{Take the [color] [shape] at [position]}
\end{itemize}

\paragraph{Feedback Expression Templates}

We use two templates to give positive or negative feedback on the direction of the follower
\begin{itemize}
    \itemsep0em
    \item \textit{Yes this way}
    \item \textit{Not this way}
\end{itemize}

And we give a similar feedback when the follower is locating the gripper over a piece
\begin{itemize}
    \itemsep0em
    \item \textit{Yes this piece}
    \item \textit{Not this piece}
\end{itemize}

\paragraph{The vocabulary}
\label{appendix:vocabulary}

 Overall, the property values and sentence templates lead to a small vocabulary of 33 words:

\begin{itemize}
    \itemsep0em
    \item 9 shapes: F, N, P, T, U, W, X, Y, Z
    \item 6 colors: red, yellow, green, blue, purple, brown
    \item 6 position words: left, right, top, bottom, center (which are combined to e.g., right center or top left)
    \item 8 template words: take, the, piece, at, yes, no, this, way
    \item 4 special words: <s>, <e>, <pad>, <unk>
\end{itemize}

The maximal sentence length is 11.

\newpage

\subsection{Follower Details}
\label{appendix:follower}

\paragraph{Agent}
Parameters: $9,456$

\begin{table}[h]
    \centering
    \begin{tabular}{| l | r| }
        \hline
         word\_embedding\_dim & 128 \\
         feature\_embedding\_dim & 128 \\
         actor\_layers & 2 \\
         actor\_dims & 128 \\
         vf\_layers & 2 \\
         vf\_dims & 128 \\
        \hline
    \end{tabular}
    \caption{Agent hyperparameters}
    \label{tab:agent_hyperparameters}
\end{table}

The max-pooling layer additionally downsamples the language conditionaed visual features from $11\times11\times128$ to $1\times1\times128$ dimensions. For this we use the \texttt{nn.AdaptiveMaxPool2d((1, 1))} layer from PyTorch v1.11.0. In addition, before we average the gripper coordinates features and the resulting language conditioned visual features, we apply a layer normalization (eps = 1e-5) on them. 

\paragraph{Architecture Search}

We performed a little architecture search where we evaluated two methods for visual encoding (pixels, symbols), four methods for language encoding (word embeddings with GRU,one-hot word embeddings with GRU, one-hot sentence embeddings, pre-trained sentence embeddings) and two methods for the fusion (concatenate, FiLM). We found learnt word embeddings and FiLM perform best in regard of training speed and success rate. The visual encodings showed similar performance but we prefer the pixel encoder because it makes less assumptions about the world.

\paragraph{Learning Algorithm}

\begin{table}[b]
    \centering
    \begin{tabular}{| l | r| }
        \hline
         lr\_init & 2.5e-4 \\
         lr\_min & 2.5e-5\\
         num\_epochs & 8\\
         buffer\_per\_env & 1024\\
         clip\_range & 0.2 \\
         clip\_range\_vf & 0.2 \\
         ent\_coef & 0.01 \\
         vf\_coef & 0.5 \\
         target\_kl & 0.015  \\
        \hline
    \end{tabular}
    \caption{PPO hyperparameters}
    \label{tab:ppo_hyperparameters}
\end{table}

We apply a learning rate schedule that decreases the learning rate during training according to the training progress (based on the number of time steps) with $p \in [0,1]$, but the learning rate is given a lower bound $\alpha_{\text{min}}$ so that it never reaches zero: $\alpha_t = \text{max}(p \cdot \alpha_{\text{init}}, \alpha_{\text{min}})$

\subsection{Environment Details}
\label{appendix:environment}

\paragraph{Board} The internal representation of the visual state is a 2-dimensional grid that spans $W \times H$ tiles where $W$ and $H$ are defined by the map size. A tile is either empty or holds an identifier for a piece (the tile is then occupied). The pieces are defined by their colour, shape and coordinates and occupy five adjacent tiles (within a virtual box of $5\times5$ tiles). The pieces are not allowed to overlap with another piece's tiles. For a higher visual variation, we also apply rotations to pieces, but we ignore the rotation for expression generation, though this could be an extension of the task. 

\begin{table}[h]
    \centering
    \begin{tabular}{|l|l|l|}
    \hline
    Name & HEX & RGB \\
    \hline
 red & \#ff0000 & (255, 0, 0) \\
 yellow & \#ffff00 & (255, 255, 0) \\
 green & \#008000 & (0, 128, 0) \\
 blue & \#0000ff & (0, 0, 255) \\
 purple & \#800080 & (128, 0, 128) \\
 brown & \#8b4513 & (139, 69, 19) \\
    \hline
    \end{tabular}
    \caption{The colors for the Pentomino pieces.}
    \label{tab:colors}
    \vspace{-0.3cm}
\end{table}

\paragraph{Gripper} The gripper can only move one position at a step and can move over pieces, but is not allowed to leave the boundaries of the board. 
The gripper coordinates $\{(x,y): x \in [0,W], y \in [0,H]\}$ are projected to $\{(x,y): x,y \in [-1,+1]\}$ so that the coordinate in the center is represented with $(0,0)$. This provides the agent with the necessary information about its positions on the overall board as its view field is shrinked to $11\times11$ tiles. 
In addition, to provide the agent with a notion of velocity, the environment keeps track of the last two gripper positions and applies a grey with decreasing intensity to these positions on the board:

\begin{itemize}
    \itemsep0em
    \item $\text{color}_{g_t}=(200,200,200)$
    \item $\text{color}_{g_{t-1}}=(150,150,150)$
    \item $\text{color}_{g_{t-2}}=(100,100,100)$
\end{itemize}

\subsection{Task Details}
\label{appendix:data_generation}

We created training, validation, test and holdout splits of target piece symbols (a combination of shape, color and position) for the task creation (see Table~\ref{table:task_sets}). We split these possible target piece symbols so that each subset still contains all colors, shapes and positions, but different combinations of them. For example, the training set might contain a ``red F`` but this is never seen at the bottom left. Though this will be seen during validation or testing. An exception is the holdout split where we hold out a color for each shape. This means that for example a ``green T'' is never seen during training, but a ``green F'' or a ``blue T''.

\begin{table}[h]
    \tiny
    \centering
    \begin{tabular}{|l|r|l|l|l|l|l|l|} 
    \hline
    \multicolumn{1}{|l}{} &       & \multicolumn{6}{c|}{\# of Tasks}                                                                                                                                   \\ 
    \cline{3-8}
    \multicolumn{1}{|l}{} &       & \multicolumn{2}{c|}{Map Size=20}                        & \multicolumn{4}{c|}{Map Size=30}                                                                                 \\ 
    \cline{2-8}
                          & \# TPS & \multicolumn{1}{c|}{N=4} & \multicolumn{1}{c|}{N=8} & \multicolumn{1}{c|}{N=4} & \multicolumn{1}{c|}{N=8} & \multicolumn{1}{c|}{N=12} & \multicolumn{1}{c|}{N=18}  \\ 
    \hline
    training              & 275   & 1650                     & 1650                     &                          &                          &                           &                            \\ 
    \cline{1-4}
    validation            & 25    & 150                      & 150                      &                          &                          &                           &                            \\ 
    \hline
    testing               & 60    & 360                      & 360                      & 180                      & 180                      & 180                       & 180                        \\ 
    \hline
    holdout               & 72    & 432                      & 432                      &                          &                          &                           &                            \\
    \hline
    \end{tabular}
    \caption{The target piece symbols (TPS) distributed over the task splits with different map sizes (Size) and number of pieces (N) on the board. The total possible number of target piece symbols is $9 \cdot 6 \cdot 8= 432$.}
    \label{table:task_sets}
    \vspace{-0.3cm}
\end{table}

To create a task we first place the target piece on a board with the wanted map size. Then we sample uniform random from all possible pieces and place them until the wanted number of pieces is reached. If a piece cannot be placed $100$ times, then we re-sample a piece and try again. The coordinates are chosen at random uniform from the coordinates that fall into an area of the symbolic description. We never set a piece into the center, because that is the location where the gripper is initially located. 

\subsection{Experiment Details}
\label{appendix:experiment}

We trained the agents on a single GeForce GTX 1080 Ti (11GB) where each of them consumed about 1GB of GPU memory. The training spanned $10.24$ million time steps executed with 4 environments simultaneously (and a batch size of $64$). The training took between $9.24$ and $12.32$ hours ($11.86$ hours on average). The random seed was set to $49184$ for all experiments. We performed an evaluation run on the validation tasks after every $50$ rollouts ($51,200$ timesteps) and saved the best performing agent according to the mean reward. 

\begin{table}[h]
    \centering
    \small
    \begin{tabular}{|c|c|c|}
    \hline
    Pr.Or. & Step in K w/ FB & Step in K w/o FB\\
    \hline
     \texttt{C-S-P} & 8,601 & 8,806\\
     \texttt{C-P-S} & 8,396 & 9,911\\
     \texttt{P-S-C} & 9,216 & 9,830\\
     \texttt{P-C-S} & 5,939 & 9,830\\
     \texttt{S-P-C} & 5,529 & 8,806\\
     \texttt{S-C-P} & 7,984 & 10,035\\
    \hline
    \end{tabular}
    \caption{The timesteps of the best model checkpoints.}
    \label{tab:checkpoints}
    \vspace{-0.3cm}
\end{table}

As the evaluation criteria on the testings tasks we chose success rate which indicates the relative number of episodes (in a rollout or in a test split) where the agent selected the correct piece: 
$$
\text{mSR}=\frac{\sum^{N}{s_i}}{N} \text{ where } s_i=\begin{cases}
    1, & \text{for correct piece} \\
    0, & \text{otherwise}
\end{cases}
$$

\newpage

\section{Additional Results}

\begin{figure}[h]
    \begin{center}
        \includegraphics[width=0.45\textwidth]{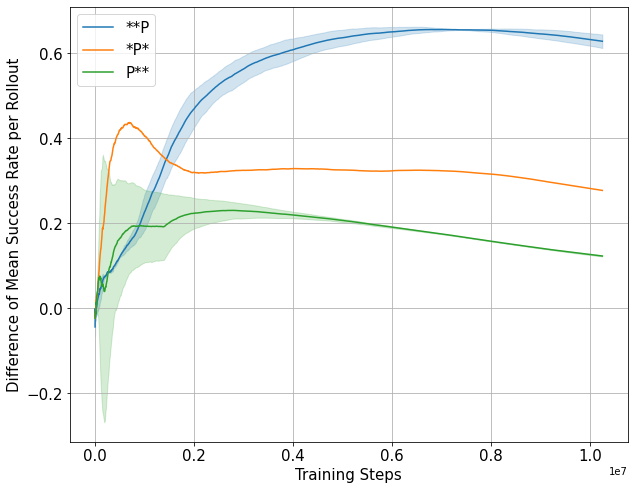}
    \end{center}
    \caption{The difference of success rates during \textit{training}, when we directly compare agents that are exposed to teacher's with the similar preference orders (\texttt{P**,*P*,**P}). The lines indicate the success rates of the agents with feedback minus the success rates of agents without feedback.}
    \label{fig:feedback}
   \vspace{-0.5cm}
\end{figure}

In addition, we notice that the feedback has a positive effect on early success rates during \textit{training} when we compare training runs of the same preference order groups with and without feedback (see \ref{fig:feedback}). The intra-episodic feedback largely improves the early success rates of agents with teachers of preference orders \texttt{**P} (\texttt{SCP}, \texttt{CSP}) as well as those with preference orders \texttt{*P*} (\texttt{SPC}, \texttt{CPS}). There is also a noticable but lower effect on the preference orders \texttt{P**} (\texttt{PSC}, \texttt{PCS}) that perform already well early without the intra-episodic feedback. Though the latter seem to be confused by the feedback initially (until 10\% of the training steps). The benefit of intra-episodic feedback is starting to decrease in later time steps, because the agent without that additional signal catch up on the success rates. Still these findings show that intra-episodic feedback is helpful to improve the learning in early stages.

\section{Misc}

Robot image in Figure~\ref{fig:example_board} adjusted from \url{https://commons.wikimedia.org/wiki/File:Cartoon_Robot.svg}.
That file was made available under the Creative Commons CC0 1.0 Universal Public Domain Dedication.

\end{document}